\documentclass[sigconf]{aamas} 

\usepackage[dvipsnames]{xcolor}
\usepackage{mathtools}
\usepackage{booktabs, array} 
\usepackage{multirow}
\usepackage[table]{xcolor}

\usepackage{balance}
\setcopyright{none}
\acmSubmissionID{aai036}

\title[MARL-GPT]{MARL-GPT:~Foundation Model for Multi-Agent Reinforcement Learning}
\subtitle{AAAI Track}

\author{Maria Nesterova}
\affiliation{
  \institution{AXXX \& MIRAI}
  \city{Moscow}
  \country{Russia}}
\email{minesterova@yandex.ru}

\author{Mikhail Kolosov}
\affiliation{
  \institution{MIRAI}
  \city{Moscow}
  \country{Russia}}
\email{mikek7532@gmail.com}

\author{Anton Andreychuk}
\affiliation{
  \institution{AXXX}
  \city{Moscow}
  \country{Russia}}
\email{andreychuk@cogailab.com}

\author{ Egor Cherepanov}
\affiliation{
  \institution{AXXX \& MIRAI}
  \city{Moscow}
  \country{Russia}}
\email{cherepanov.e@miriai.org}

\author{Oleg Bulichev}
\affiliation{
  \institution{MIRAI \& Innopolis University}
  \city{Moscow}
  \country{Russia}}
\email{o.bulichev@innopolis.ru}

\author{Alexey Kovalev}
\affiliation{
  \institution{AXXX \&  MIRAI}
  \city{Moscow}
  \country{Russia}}
\email{	alexeykkov@gmail.com}

\author{Konstantin Yakovlev}
\affiliation{
  \institution{SPbU \& AXXX}
  \city{Saint-Petersburg}
  \country{Russia}}
\email{k.yakovlev@spbu.ru}

\author{Aleksandr Panov}
\affiliation{
  \institution{AXXX  \&  MIRAI}
  \city{Moscow}
  \country{Russia}}
\email{panov@axxx.tech}

\author{Alexey Skrynnik}
\affiliation{
  \institution{AXXX  \&  MIRAI}
  \city{Moscow}
  \country{Russia}}
\email{skrynnikalexey@gmail.com}

\begin{abstract}
Recent advances in multi-agent reinforcement learning (MARL) have demonstrated success in numerous challenging domains and environments, but typically require specialized models for each task. 
In this work, we propose a coherent methodology that makes it possible for a single GPT-based model to learn and perform well across diverse MARL environments and tasks, including StarCraft Multi-Agent Challenge, Google Research Football and POGEMA. 
Our method, MARL-GPT, applies offline reinforcement learning to train at scale on the expert trajectories (400M for SMACv2, 100M for GRF, and 1B for POGEMA) combined with a single transformer-based observation encoder that requires no task-specific tuning.
Experiments show that MARL-GPT\footnote{Code: \url{https://github.com/Cognitive-AI-Systems/marl-gpt}} achieves competitive performance compared to specialized baselines in all tested environments. 
Thus, our findings suggest that it is, indeed, possible to build a multi-task transformer-based model for a wide variety of (significantly different) multi-agent problems paving the way to the fundamental MARL model (akin to ChatGPT, Llama, Mistral etc. in natural language modeling).

\end{abstract}

         
\newcommand{\BibTeX}{\rm B\kern-.05em{\sc i\kern-.025em b}\kern-.08em\TeX}
\newcommand{\rot}[1]{\rotatebox[origin=c]{60}{#1}}

\begin{document}

\pagestyle{fancy}
\fancyhead{}

\maketitle 

\section{Introduction}

\begin{figure}[t!]
    \centering
    \includegraphics[width=1.0\linewidth]{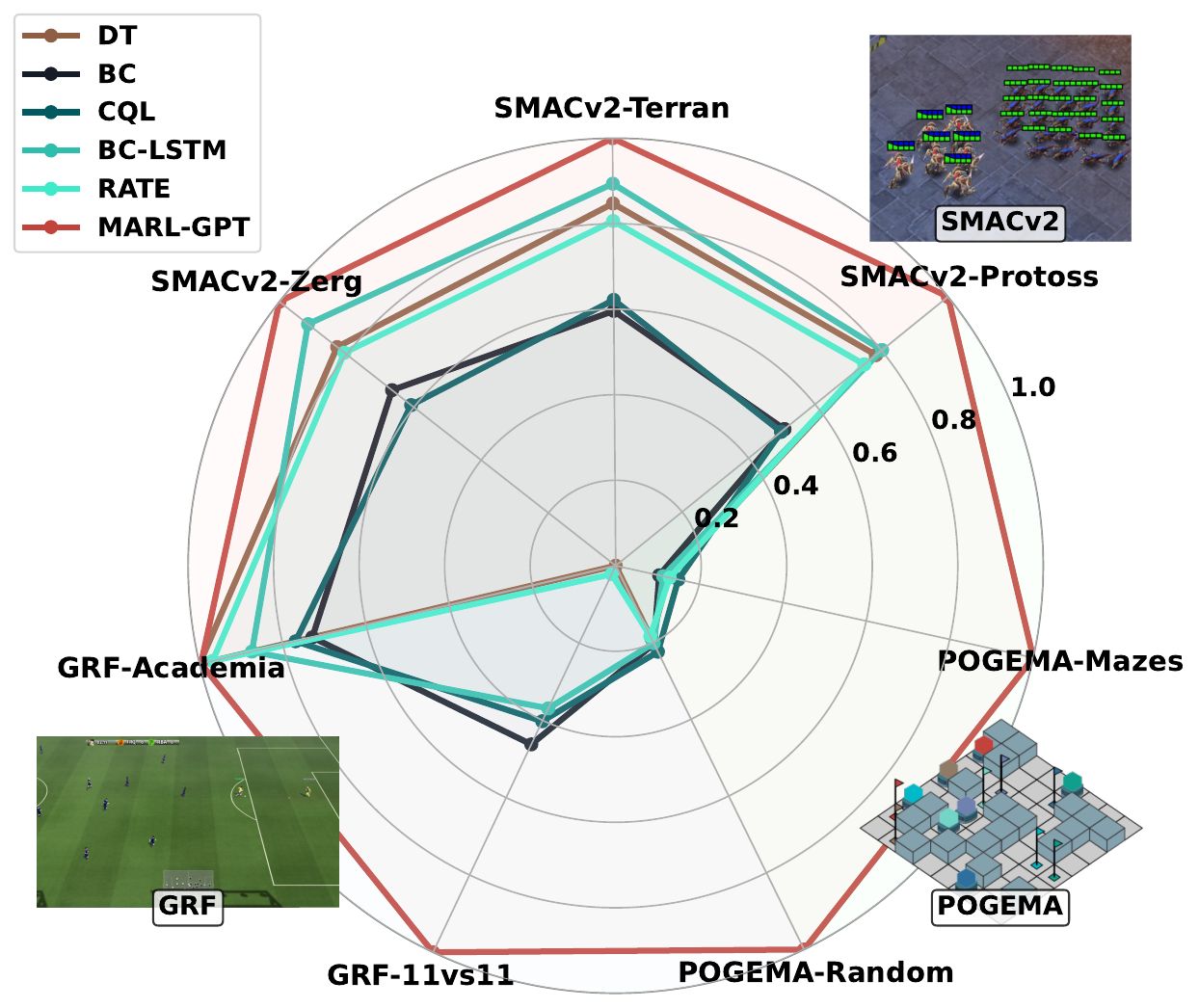}
    \caption{Spider plot demonstrating the relative performance of all evaluated approaches on three different environments -- SMACv2, Google Research Football (GRF) and POGEMA.}
    \Description{Spider plot demonstrating the relative performance of all evaluated approaches on three different environments -- SMACv2, Google Research Football (GRF) and POGEMA.}
    \label{fig:spider_plot}
\end{figure}

Multi-agent reinforcement learning (MARL) has made significant progress in solving complex tasks such as competitive games like StarCraft or cooperative robotic tasks such as multi-agent pathfinding. However, most MARL methods are designed for a single environment or task, requiring specialized architectures and training pipelines for each new problem. And even in this case the resultant MARL policies typically struggle when solving the problems not seen in the training time that are different in the number of agents, map layout etc. In this work, we wish not only to improve the generalization abilities of a MARL policy but to make it possible for such policy to efficiently solve tasks from different domains.

\begin{figure*}
    \centering
    \includegraphics[width=0.95\linewidth]{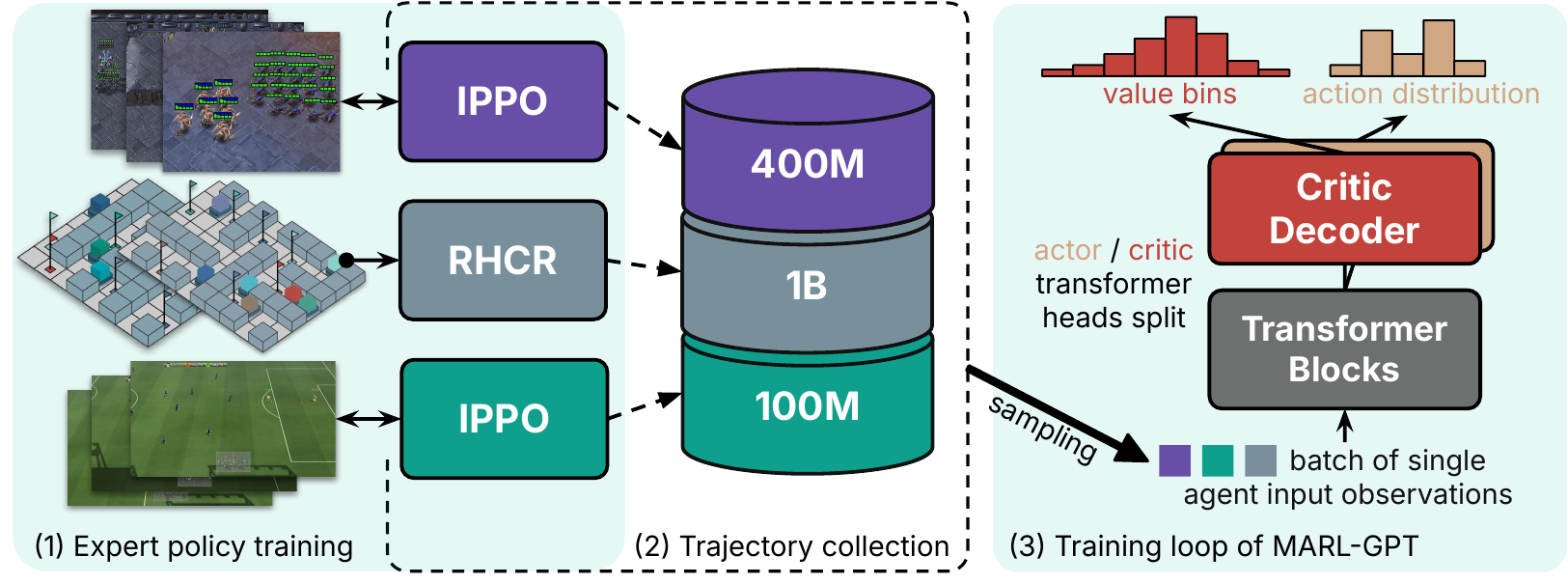}
    \caption{The general pipeline begins with training expert policies across diverse MARL environments using domain-appropriate algorithms (e.g., IPPO, RHCR). These policies generate large-scale datasets of observation-action(-reward) trajectories. MARL-GPT is then trained on the aggregated data using cross-entropy loss to predict expert actions and Q-values, enabling a single model to generalize across environments, tasks, and coordination regimes.}
    \Description{The general pipeline begins with training expert policies across diverse MARL environments using domain-appropriate algorithms (e.g., IPPO, RHCR). These policies generate large-scale datasets of observation-action(-reward) trajectories. MARL-GPT is then trained on the aggregated data using cross-entropy loss to predict expert actions and Q-values, enabling a single model to generalize across environments, tasks, and coordination regimes.}
    \label{fig:enter-label}
\end{figure*}

To this end we leverage the power of transformer-based architectures (that are the backbone of the overwhelming success in natural language processing tasks and are successfully applied to other domains such as computer vision, robotic control as well) coupled with imitation-learning at scale. We focus on three significantly different MARL environments: Starcraft Multi-Agent Challenge (adversarial combat game)~\citep{samvelyan2019starcraft}, Google Research Football (adversarial sport game)~\citep{kurach2020google}, and POGEMA (cooperative multi-robot navigation)~\citep{skrynnik2025pogema}. Using the expert policies for those environments that include both learnable policies and the rule-based ones, we obtain a large and diverse dataset of observation-action pairs that are used to train our model, which we call MARL-GPT.

We train MARL-GPT using supervised imitation learning on a diverse dataset of expert demonstrations collected from multiple MARL environments and tasks (Fig.~\ref{fig:enter-label}). Each expert trajectory consists of sequences of observation-action-reward triplets gathered across scenarios that vary in dynamics, state spaces, and agent interactions. To represent these observations, we tokenize structured, vectorized inputs describing each agent’s local view -- including itself, allies, and opponents. These tokens are enriched with the learned embeddings that encode the type of feature (e.g., position, health), the agent’s identity and team affiliation, and the temporal step of the observation. This flexible encoding supports a variable number of agents and preserves role-specific and group-specific information while maintaining permutation invariance where applicable. The resulting tokens are processed by a transformer encoder, which maps them to expert action predictions and Q-values. Training is performed using standard cross-entropy loss on discrete action outputs or Q-value bins, without relying on recurrence or autoregression. Instead, partial observability is handled by leveraging the transformer’s context window and temporal encoding.

As a result we are able to obtain a single (fundamental) MARL model, that generalizes across state and action spaces, reward structures, and agent coordination requirements as confirmed by our extensive empirical evaluation.

Our experiments show that this unified model achieves competitive performance compared to specialized MARL algorithms in all tested domains. This suggests that general-purpose architectures can handle diverse multi-agent tasks without extensive tuning, reducing the need for task-specific designs.

In summary, our key contributions are as follows:
\begin{enumerate}
    \item We develop a single GPT-based MARL model that performs notably well across different environments without any architectural changes or fine-tuning. 
    \item We create a large dataset of observation-action-reward triplets, vital for imitation learning and offline RL of any MARL policy (not necessarily ours).
    \item We conduct a thorough empirical validation of our model on SMACv2, Google Research Football, and POGEMA, showing competitive results against specialized baselines.
\end{enumerate}

We are committed to open-sourcing the model code, including expert policy weights, training datasets, and final weights of the MARL-GPT model.

\section{Related Work}

\paragraph{Multi-agent imitation learning (MAIL)} In the realm of multi-agent systems, imitation learning and learning from demonstration are widely employed~\citep{tang2024multiagent,liu2024learning}. 
Imitation learning in multi-agent scenarios, also known as MAIL, is a problem in which agents learn to perform a task in a multi-agent system by observing and mimicking expert demonstrations without any knowledge of the reward function from the environment. This approach has gained particular traction in the context of controlling urban traffic and traffic lights at intersections~\citep{bhattacharyya2018multi,huang2023multi}, controlling the power in wireless networks~\citep{zhang2025multi} due to the availability of a vast amount of data collected in real-world scenarios and the use of high-quality simulators, e.g. Sumo~\citep{lopez2018microscopic} in traffic applications. 

Within the realm of MAIL, there are various methods to consider, including those that employ Bayesian approaches~\citep{yang2020bayesian}, generative adversarial techniques~\citep{song2018multi,li2024gailpg}, statistical tools for capturing interdependencies between agents~\citep{wang2021multi}, low-rank subspaces~\citep{shih2022conditional}, latent models for coordinating agents~\citep{le2017coordinated}, decision transformers~\citep{meng2023offline,wen2022multi}, and more. Demonstrations are frequently used for pretraining in games, such as learnable models for chess~\citep{silver2016mastering, ruoss2024amortized}, and in multi-agent pathfinding tasks, as exemplified by MAPF-GPT~\citep{andreychuk2025mapf} and SCRIMP~\citep{wang2023scrimp}. Despite the presence of models that are trained on pre-collected data, they are often specialized for a specific environment or even a single task within that environment. These models have limited generalizability even to conditions within the same environment, and their performance is significantly reduced when learning in a multitasking setting.

\paragraph{Foundation models for multi-agent systems.} 
Foundation models are typically trained on large datasets, enabling zero-shot or few-shot learning~\citep{bommasani2021opportunities,yang2023foundation}. 
From the perspective of an autonomous agent, it is a model that can perform new tasks that are different from those it was trained on, either with additional demonstration of desired behavior or without any demonstration~\citep{firoozi2023foundation}. Another important feature of these foundation models is their fine-tuning ability for specific tasks, allowing them to improve performance quickly (see Gato tuning~\citep{reed2022a}). These models are commonly used in robotics to solve multimodal tasks where the goal is specified in text instructions~\citep{firoozi2023foundation,team2024octo,kim2024openvla}. However, in multi-agent scenarios, such models are less common, with examples including the Magnetic-One model for language and multimodal task-solving in WebArena~\citep{fourney2024magentic}, as well as the MAPF-GPT model for pathfinding~\citep{andreychuk2025mapf}. 

Work on multitasking multi-agent models is also close to this topic, in which attempts are made to train a single policy for action in different scenarios and environments~\citep{park2025spectra}.  The primary direction for developing foundation models in the multi-agent setting is based on transformer architectures. First, transformers can process observations of varying lengths~\citep{hu2021updet}, which is particularly important when the dimensionality of the observation depends on the number of agents in the task. Second, transformers enable the model to capture inter-agent relationships, which is valuable not only for observation processing, but also for partially centralized training approaches~\citep{wen2022multi}. Additionally, several offline methods~\citep{zhang2022discovering,liu2025learning} extract coordination skills that are generalized between grouping agents. However, these approaches lack the key properties of foundation models, particularly the ability to fine-tune on additional demonstrations, and they struggle to effectively train in multi-task settings with distinct environments.

Some papers have explored the possibility of adapting single-agent foundation models to solve multi-agent problems without modifying the pre-training process~\citep{veerapaneni2024work,xu2024multi}. 
In our method, we focus on the task of training a foundation model for multiple multi-agent environments from scratch. It should be noted that no such single pre-trained model has been proposed yet. This is due to the complex nature of multi-agent policies in various tasks, such as StarCraft~\citep{samvelyan2019starcraft} and football~\cite{kurach2020google}, and the lack of large expert trajectory datasets necessary for effective foundation model training. In our MARL-GPT model, we overcome these difficulties and show that it is possible to train a multitasking, multi-agent foundation model on a reasonable set of expert data and with expert-level quality.

\section{Background}
{
\textbf{Multi-Agent Reinforcement Learning} extends single-agent reinforcement learning (RL) to domains with multiple interacting decision makers. In partially observable environments, this interaction can be formalized by a decentralized partially observable Markov decision process (Dec-POMDP)~\cite{oliehoek2016concise}. This is a tuple
$(N, \mathcal{S}, \mathcal{A}, T, \{\mathcal{O}^i\}_{i \in N}, O, \{\mathcal{R}^i\}_{i \in N}, \gamma),$
where $N$ is the agent set, $\mathcal{S}$ is the global state space, and all agents share a common action space $\mathcal{A}$. The joint action is $\mathbf{a} = (a^1, \dots, a^n) \in \mathcal{A}^n$. The transition kernel $T(s' \mid s, \mathbf{a})$ specifies the probability of moving from state $s$ to $s'$ given the joint action. Each agent receives a private observation $o^i \in \mathcal{O}^i$ drawn from the observation function $O(o^1, \dots, o^n \mid s, \mathbf{a}, s')$. Rewards may differ across agents, with $\mathcal{R}^i(s, \mathbf{a}, s')$ denoting the signal for agent $i$, and $\gamma \in [0, 1)$ the discount factor.}

{At each time step $t$, agent $i$ selects its action based on its own interaction history via a policy $\pi^i : (o_0^i, a_0^i, \dots, o_t^i) \mapsto \Delta(\mathcal{A}),$ where $\Delta(\cdot)$ denotes the distribution of actions provided by the policy.
After all agents act, the environment transitions and each agent receives $r_t^i = \mathcal{R}^i(s_t, \mathbf{a}_t, s_{t+1}).$
The learning objective in MARL is for each agent $i$ to find an optimal policy $\pi^{*~i}$ that maximizes its expected cumulative return, which is a discounted sum of step-wise rewards.}

Each agent deploys its learned policy, which selects actions based only on its own local observation history. The policies themselves are conditioned only on local information, making execution fully decentralized. Agents do not need to communicate or observe others' states/actions at runtime.

Parameter sharing employs a single policy network $\pi$ across all agents. This approach reduces the parameter count and enables efficient knowledge transfer, though it may limit policy diversity when agents require specialized behaviors. 

Offline RL and imitation learning methods train on expert trajectories $\mathcal{D}$. The simplest approach is behavior cloning~\cite{hussein2017imitation}, which formulates imitation learning as supervised learning by minimizing the cross-entropy loss $\mathbb{E}_{o_t, a_t  \sim \mathcal{D}} \left[\mathcal{L}_{CE}( a \sim \pi(a | o_t), a_t)\right]$ over expert demonstrations $(o_t, a_t)$.

\section{Method}

The suggested pipeline of obtaining a multi-task MARL model can be divided into three main components. First, expert policies for each task are acquired, and expert trajectories are collected using them. Second, observations are encoded in a way that enables the model to operate across multiple environments, in contrast to the expert policies, which are specialized for a single environment. Finally, a transformer architecture is trained via behavior cloning to imitate the expert behavior.

\subsection{Acquiring of Expert Policies and Data Collection}

Let $\pi^*_e$ denote an expert policy for the environment $e \in \mathcal{E}$. These expert policies serve as sources of high-quality demonstrations and can be obtained through a variety of means. In certain well-established domains, such as chess, expert policies are readily available in the form of rule-based or search-based engines (e.g., Stockfish with NNUE), which do not require further training. These systems can generate large volumes of expert trajectories with negligible additional cost.

In contrast, expert policies for complex multi-agent systems are significantly less common. One exception is the domain of multi-agent pathfinding, where centralized solvers (e.g., Conflict-Based Search) can be used to compute globally optimal or near-optimal solutions. These solvers are environment-specific but provide valuable expert supervision when applicable.

In many modern MARL environments, no public expert policies are readily available. In such cases, we construct strong policies by applying large-scale reinforcement learning. For instance, we use independent PPO agents trained separately in each environment $e$ to convergence, resulting in a set of high-performing task-specific policies. This approach is applicable even in partially observable settings.

Each expert policy $\pi^*_e$ is used to collect trajectories by interacting with its corresponding environment. If the policy is recurrent, it may condition on a history of past observations. Specifically, the expert produces a trajectory:
\[
\tau_e = \{(o_t, a_t, r_t)\}_{t=1}^T,
\]
where $o_t$ is the observation at time $t$, and $a_t \sim \pi^*_e(\cdot \mid h_t)$ is the action sampled based on observation history $h_t = \{o_{t-k}, \dots, o_t\}$. The agent receives the reward $r_t$. In practice, the policy may use a GRU or similar recurrent architecture to encode $h_t$ into a compact hidden state.  

We collect expert trajectories across a distribution of environments $\mathcal{E}$. Some of them may contain multiple different scenarios (tasks), allowing us to gather data that vary not only across environments, but also across tasks within a given environment. This results in a diverse dataset of expert demonstrations covering a wide range of state spaces, dynamics, and task structures. 

\subsection{Observation Encoding}

We consider the task of multi-agent decision making in a partially observable environment. In this setting, observations usually include information about the agent itself, the surrounding agents, and the environment around them. We assume that observation from any environment can be represented as a vector $o = \{o_i\}_{i = 0}^n$. For each element $o_i$ in this observation vector, we obtain a token embedding $tok_i$ by applying a learned linear layer $tok_i = L(o_i)$ that projects the raw input into the embedding space. $L$ is the same for all environments and maps a number into a vector.

Each element of the observation corresponds to a specific environmental property, e.g. the health of the nearest agent. To provide context to these elements, we augment them with positional and structural information. Four positional numbers are defined to encode this information. First, $pos_{attr}$ (attribute) indicates the type of observed property. For instance, whether it is the agent's health or its coordinates. If both agent $k$ and agent $j$ have the same property $P$ observed, they will share the same $pos_{attr}$. Second, $pos_{team}$ denotes the group to which the observed agent belongs. Agents may be divided into different groups based on their roles or goals, such as allies and enemies. Third, agents can be numbered within their group using $pos_{indx}$ . Thus, an observation element $o_i$ can be associated with an agent of the group $pos_{team}$ and the index $pos_{indx}$. The agent whose observation is being considered is always assigned group $0$ and index $0$. All global properties of the situation are also attributed to this agent (group $0$, index $0$).

\begin{figure}[ht!]
    \centering
    \includegraphics[width=1.0\linewidth]{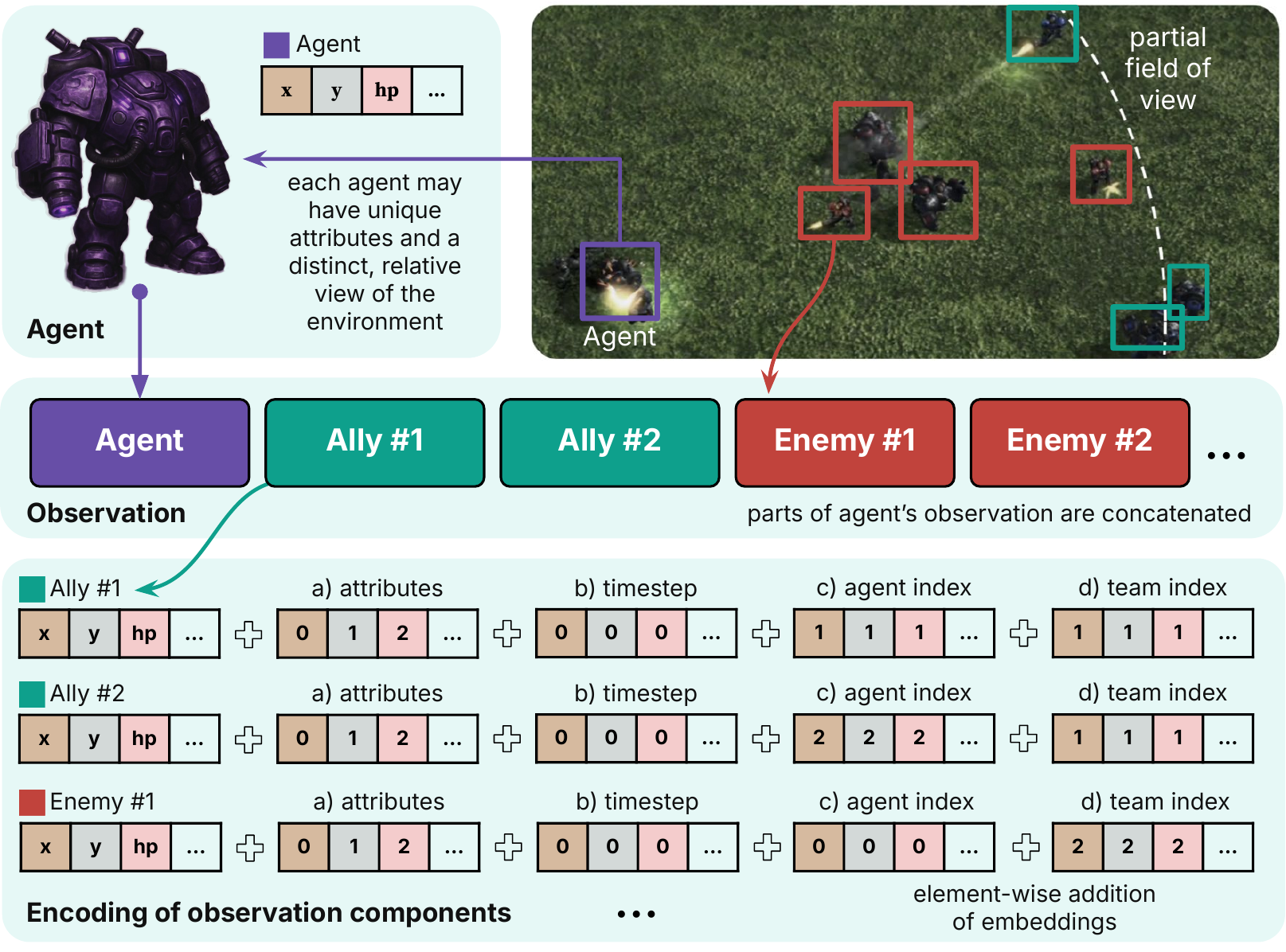}
    \caption{Illustration of the proposed encoding scheme for multi-agent systems, demonstrated using an example from the SMACv2 environment.  Each agent receives a structured, vectorized observation containing information about itself, its allies, and nearby enemies. For each observed agent, the input features (such as position, health, etc.) are enriched with additional embeddings: (a) a positional encoding over the feature dimensions, (b) a global timestep encoding, (c) an agent identity index, and (d) a team index distinguishing allies from enemies. These components are combined via element-wise addition, resulting in a contextualized embedding for each observed agent that can be processed by a transformer encoder. The provided encoding scheme is general and transferable across multi-agent systems. By augmenting raw observation features with agent- and team-specific indices, the encoding enables the model to distinguish between different functional groups (e.g., allies vs. enemies) and individual agents, regardless of environment-specific dynamics or layouts. This structure preserves permutation invariance where appropriate, while still allowing the model to learn role-specific interactions. The encoding supports a dynamic number of agents, limited only by the model’s maximum context size.}
    \Description{Illustration of the proposed encoding scheme for multi-agent systems, demonstrated using an example from the SMACv2 environment.  Each agent receives a structured, vectorized observation containing information about itself, its allies, and nearby enemies. For each observed agent, the input features (such as position, health, etc.) are enriched with additional embeddings: (a) a positional encoding over the feature dimensions, (b) a global timestep encoding, (c) an agent identity index, and (d) a team index distinguishing allies from enemies. These components are combined via element-wise addition, resulting in a contextualized embedding for each observed agent that can be processed by a transformer encoder. The provided encoding scheme is general and transferable across multi-agent systems. By augmenting raw observation features with agent- and team-specific indices, the encoding enables the model to distinguish between different functional groups (e.g., allies vs. enemies) and individual agents, regardless of environment-specific dynamics or layouts. This structure preserves permutation invariance where appropriate, while still allowing the model to learn role-specific interactions. The encoding supports a dynamic number of agents, limited only by the model’s maximum context size.}
    \label{fig:pos-example}
\end{figure}

The environment is partially observable, which means that the agent does not have access to the full state of the environment. One simple way to handle this limitation is to include in the final observation a history of the last $h$ observations from the environment, or to complete the current observation with some information from previous steps. To indicate the time moment of each observation element, we add a positional encoding $pos_{time}$. It is important to note that we do not use methods such as autoregressive modeling or recurrent neural networks in this approach.

Thus, each observation element $o_i$ can be represented by a tuple of four components: agent index $pos_{indx}^i$, group number $pos_{team}^i$, agent property number $pos_{attr}^i$, and time step $pos_{time}^i$ from which the observation was taken. The structure of these positional vectors depends on the specific environment. For each positional component, a corresponding embedding is learned $emb_{type}^i = L_{type}(pos_{type}^i)$ and added to the respective token.

In summary, each observation element $o_i$ is contextualized by a four-component positional vector: the agent index $pos_{indx}^i$, group identifier $pos_{team}^i$, attribute type $pos_{attr}^i$, and timestep $pos_{time}^i$. While the rules for assigning these positional indices are environment-dependent, the mechanism for using them is universal. Each positional component is mapped to a vector using a learned embedding layer $emb_{type}^i = L_{type}(pos_{type}^i)$. These four positional embeddings are then summed with the initial observation token $tok_i$ to produce the final input representation, $res_i$.
\begin{equation*}
    res_i = tok_i + emb_{indx}^i + emb_{team}^i + emb_{attr}^i + emb_{time}^i.
\end{equation*}

The result tokens $res_i$ are then processed by a GPT-like model. The example of a positional encoding for the well-known SMACv2 environment is shown in Fig.~\ref{fig:pos-example}. Appendix~A provides recommendations for constructing positional encodings tailored to specific environment.

\subsection{Model Training}

We study a universal model that predicts an agent’s action using only its own observation, without access to the state or observations of other agents. Our approach is based on offline RL with a data-driven actor-critic architecture. 
At each timestep $t$, the transformer-based observation encoder processes the agent’s observation $o_t$ into a hidden state $h_t$, which is shared by both the critic $Q$ and stochastic policy $\pi$ output heads. The model is trained by minimizing joint loss $\mathcal{L} = \mathcal{L}_{{critic}} + \mathcal{L}_{{actor}}$ using minibatch stochastic gradient descent with the Adam optimizer. 

\textbf{Critic.}
Classification with categorical cross-entropy loss ($\mathcal{L}_{CE}$) is more effective than regression with mean squared error when training large neural networks, especially transformer architectures. For this reason, we use a discrete critic approach~\cite{ruoss2024amortized, farebrother2024stop}. The Q-value range is divided into a fixed number of uniform bins, and each Q-value is represented as a probability distribution over these bins. If $K$ is the scalar, then $K^B$ represents this scalar as a vector of bins.

The critic loss consists of two parts: the temporal difference (TD) error and the conservative regularization (CR). The TD target is $y_t = r_t + \gamma \; \bar{Q}(o_{t+1}, a \sim \pi(a | o_t))$, where $\bar{Q}$ and $\bar{\pi}$ refer to frozen target networks. The regularization term addresses the issue of overestimating Q-values for observation-action pairs that are rarely or never seen in the dataset~\cite{kumar2020conservative,chebotar2023q}. To prevent this, we penalize the Q-values for actions with low probability under the current policy, pushing them toward a minimal attainable value $q_{\text{min}}$. Specifically, we define $\pi'(a \mid o) = \frac{1 - \pi(a \mid o)}{Z}$ as a distribution over actions that have a low probability according to $\pi$, where $Z$ is a normalization constant. The critic loss is:
\begin{multline}
    \mathcal{L}_{critic} = \alpha_{td} \; \mathbb{E}_{o, a \sim \mathcal{D}} \left[ \mathcal{L}_{CE}( Q^B(o, a), y^B_t)\right] + \\ + \alpha_{cr} \; \mathbb{E}_{o \sim \mathcal{D}, \; a \sim \pi'(a | o)} \left[ \mathcal{L}_{CE}( Q^B(o, a), q^B_{min})\right].
\end{multline}

\textbf{Policy.} The policy loss combines behavior cloning (BC) loss and actor loss (A). We calculate actor loss using the advantage function $A(o_t, a_t)$ because it offers a stable and scale-invariant learning signal. This signal helps the actor focus on actions that perform better than the average, rather than just their absolute value. 
$$A(o_t, a_t) = Q(o_t, a_t) - \mathbb{E}_{a \sim \pi(a | o_t)}\left[ Q(o_t, a)\right]$$
We include a behavior cloning loss in the policy training because it guides the model to imitate the best behavior observed in the dataset. This loss provides a strong and reliable supervisory signal, especially when the optimal policy is unknown or difficult to estimate. The policy loss is:
\begin{multline}
    \mathcal{L}_{actor} = - \alpha_{a} \; \mathbb{E}_{o_t, a_t  \sim \mathcal{D}} \left[A(o_t, a_t) \log \pi (a_t | o_t)\right] \\ + \alpha_{bc} \; \mathbb{E}_{o_t, a_t  \sim \mathcal{D}, \; a \sim \pi(a | o_t)} \left[\mathcal{L}_{CE}( a, a_t)\right].
\end{multline}

\textbf{Action Space.} {To handle varying action spaces across environments, a universal output layer (one transformer layer plus a linear layer) is used, combined with environment-specific action masking. This masking ensures that the probability of unavailable actions is always zero, allowing the agent to adapt to different scenarios while maintaining a consistent architecture.}

\textbf{Unified model for all environments.} The model architecture and all training hyperparameters are identical across environments. Only the positional encoding and the action mask are environment dependent.

\textbf{{Online fine-tune.}}
{During offline training, the agent learns mainly from successful trajectories but tends to underestimate Q-values for unseen observations and actions. This leads to incorrect Q-value estimates during the online phase. To mitigate this, the actor is frozen for the initial $n$ steps of online training while the critic is pretrained using online interactions with the environment. Afterward, online fine-tuning is performed using PPO, which helps stabilize learning and adapt the policy safely to new data.}

\section{Empirical Evaluation}
\subsection{Experimental Setup}

\definecolor{turq}{HTML}{0ea08b}
\newcolumntype{C}{>{\centering\arraybackslash}p{2.1em}}
\newcommand{\heat}[2]{\cellcolor{turq!#1!white}{#2}}

\begin{table*}[hbt!]
\centering
\small
    \resizebox{\linewidth}{!}{
\begin{tabular}{llcccccccc}
\toprule
& & \multicolumn{5}{c}{\textbf{Single-Environment Baselines}} & \textbf{Multi-Env} & \textbf{Single-Task} \\
\cmidrule(lr){3-7} \cmidrule(lr){8-8} \cmidrule(lr){9-9}
\textbf{Environment} & \textbf{Task} & DT & BC & CQL & BC-LSTM & RATE & MARL-GPT & Expert \\
\midrule

\multirow{6}{*}{SMACv2} 
& protoss 5\_vs\_5   & 82 $\pm$ 3 & 61 $\pm$ 3 & 57 $\pm$ 3 & 85 $\pm$ 3 & 79 $\pm$ 3 & \heat{30}{89 $\pm$ 3} & 87 $\pm$ 3 \\
& protoss 5\_vs\_6   & 30 $\pm$ 4 & 12 $\pm$ 4 & 14 $\pm$ 4 & 30 $\pm$ 4 & 28 $\pm$ 4 & \heat{30}{54 $\pm$ 4} & 49 $\pm$ 4 \\
& terran 5\_vs\_5    & 84 $\pm$ 3 & 69 $\pm$ 3 & 69 $\pm$ 3 & 88 $\pm$ 3 & 85 $\pm$ 3 & \heat{30}{93 $\pm$ 2} & 91 $\pm$ 2 \\
& terran 5\_vs\_6    & 48 $\pm$ 4 & 24 $\pm$ 4 & 28 $\pm$ 4 & 51 $\pm$ 4 & 41 $\pm$ 4 & \heat{30}{63 $\pm$ 4} & 61 $\pm$ 4 \\
& zerg 5\_vs\_5      & 65 $\pm$ 3 & 56 $\pm$ 3 & 50 $\pm$ 3 & 72 $\pm$ 3 & 64 $\pm$ 3 & \heat{30}{74 $\pm$ 3} & 72 $\pm$ 3 \\
& zerg 5\_vs\_6      & 34 $\pm$ 4 & 24 $\pm$ 4 & 23 $\pm$ 4 & 38 $\pm$ 4 & 33 $\pm$ 4 & \heat{30}{46 $\pm$ 4} & 48 $\pm$ 4 \\

\midrule

\multirow{6}{*}{GRF} 
& pass and shoot     & 80 $\pm$ 2 & 44 $\pm$ 2 & 60 $\pm$ 2 & 90 $\pm$ 2 & 78 $\pm$ 2 & \heat{30}{96 $\pm$ 2} & 97 $\pm$ 2 \\
& corner             & \heat{30}{60 $\pm$ 5} & 37 $\pm$ 4 & 30 $\pm$ 4 & 22 $\pm$ 4 & 58 $\pm$ 5 & 43 $\pm$ 5 & 40 $\pm$ 4 \\
& counterattack easy & 88 $\pm$ 3 & 86 $\pm$ 3 & 87 $\pm$ 3 & 88 $\pm$ 3 & 85 $\pm$ 3 & \heat{30}{89 $\pm$ 3} & 89 $\pm$ 3 \\
& 11 vs 11 easy      &  0  {$\pm$ 0}  & 40  {$\pm$ 3} & 38  {$\pm$ 3} & 43  {$\pm$ 4} &  4  {$\pm$ 3} & \heat{30}{98  {$\pm$ 2}} & 99  {$\pm$ 1}\\
& 11 vs 11 medium    &  0   {$\pm$ 0} & 41  {$\pm$ 3} & 35  {$\pm$ 4} & 30  {$\pm$ 3} &  1  {$\pm$ 1} & \heat{30}{98  {$\pm$ 3}} & 100 {$\pm$ 0} \\
& 11 vs 11 hard      &  0   {$\pm$ 0} & 40  {$\pm$ 3} & 34  {$\pm$ 4} & 24  {$\pm$ 5} &  1  {$\pm$ 1} & \heat{30}{68  {$\pm$ 7}} & 94  {$\pm$ 1}\\

\midrule

\multirow{4}{*}{POGEMA} 
& Random       & 0.22 $\pm$ 0.01 & 0.24 $\pm$ 0.01 & 0.26 $\pm$ 0.01 & 0.23 $\pm$ 0.01 & 0.21 $\pm$ 0.01 & \heat{30}{1.16  $\pm$ 0.04} & 2.16 $\pm$ 0.13 \\
& Mazes        & 0.12 $\pm$ 0.01 & 0.10 $\pm$ 0.01 & 0.14 $\pm$ 0.01 & 0.11 $\pm$ 0.01 & 0.12 $\pm$ 0.01 & \heat{30}{0.96  $\pm$ 0.04} & 1.55 $\pm$ 0.08 \\
& Warehouse    & 0.13 $\pm$ 0.01 & 0.14 $\pm$ 0.01 & 0.17 $\pm$ 0.01 & 0.12 $\pm$ 0.01 & 0.13 $\pm$ 0.01 & \heat{30}{1.02  $\pm$ 0.01} &  \\
& Cities-tiles & 0.68 $\pm$ 0.03 & 0.46 $\pm$ 0.02 & 0.66 $\pm$ 0.03 & 0.49 $\pm$ 0.02 & 0.65 $\pm$ 0.03 & \heat{30}{2.72  $\pm$ 0.12} &  \\

\bottomrule
\end{tabular}
}
\caption{Win rates \% (for SMACv2 and GRF) and average throughput (for POGEMA). Higher is better. The table compares MARL-GPT (Multi-Domain) to expert single-task policies and standard offline RL baselines (Single-Domain) across SMACv2, GRF, and POGEMA. MARL-GPT shows strong generalization, often matching or outperforming the expert and baselines. The highlighting indicates the best-performing model (excluding single-task experts).}
\label{tab:results-multitask}
\vspace{-20px}
\end{table*}

\subsubsection{Environments.}
To evaluate MARL-GPT, we selected three well-known multi-agent environments. SMACv2~\cite{smacv2} is a challenging real-time strategy benchmark that tests coordinated decision-making and teamwork in complex combat. GRF (Google Research Football)~\cite{kurach2020google} provides a dynamic, stochastic setting with varied strategies, assessing policy adaptability and temporal reasoning. The POGEMA benchmark\footnote{{https://github.com/Cognitive-AI-Systems/pogema-benchmark}}~\cite{skrynnik2025pogema} involves multi-agent pathfinding in grid worlds, providing a tough setting for testing scalable generalization to new agent populations and scenarios. We use the lifelong scenario, where a new goal is generated when an agent completes previous one.

\subsubsection{Expert Policy Generation.} 
To prepare the expert dataset for SMACv2 and GRF, we used large-scale asynchronous IPPO\footnote{https://github.com/alex-petrenko/sample-factory}. In each SMACv2 training environment, IPPO was trained using a total of 1 billion collected observations across scenarios involving \textit{Protoss}, \textit{Terran}, and \textit{Zerg}, with agent populations in both balanced (5vs5, 10vs10) and imbalanced (5vs6) settings. The resulting policy was then used to obtain 400 million agent training samples. We trained expert policies for each GRF scenario using IPPO. For the full 11vs11 match, we trained three distinct expert policies, one for each opponent difficulty level: easy, medium, and hard. Each of these policies was trained for 20 billion environment steps. For the three standard Academy scenarios, \textit{Pass and Shoot with Keeper} ran for 300 million timesteps, \textit{Easy Counter‐attack} for 400 million timesteps, and \textit{Corner} for 200 million timesteps. For the POGEMA environment, we used a centralized solver called RHCR~\cite{li2021lifelong}, in contrast to IPPO, which serves as the expert policy in other settings. RHCR operates by decomposing lifelong MAPF into windowed planning instances, resolving collisions within bounded time horizons while having access to the full state of the environment. The performance of such a centralized policy serves as an upper bound for any learnable policy, which does not have access to the full state.

\subsubsection{Trajectory Collection.}
We then used these pre-trained experts to generate multi‐agent trajectories. At every timestep in each scenario, we recorded the observations $o_t$ (as floats), the expert actions $a_t$, available actions $m_t$, scalar rewards $r_t$, and done flags $d_t$ for each agent. The tuple $(o_t, a_t, m_t, r_t, d_t)$ represents a single element in the dataset. For each SMACv2 task, we collected 60 million data elements for tasks with 5 agents and 80 million for those with 10 agents. For GRF this produced approximately 200,000 trajectories from \textit{Pass and Shoot with Keeper}, 25,000 from \textit{Easy Counter‐attack}, 75,000 from \textit{Corner}, and 30,000 from the full 11vs11 match. For POGEMA, we collected 1 billion elements in mazes and 120 million on random maps, using 32 agents and 256 steps per episode. We designed a custom dense reward for POGEMA: agents receive 1 when moving toward the goal, and 0 otherwise.

\subsubsection{Training.}
All versions of MARL-GPT were trained using 7M parameters, a history window of 6 {(except POGEMA)}, and the full dataset (except SMACv2 tasks with 10 agents). The main model was trained for 500,000 timesteps, while models used for testing theories were trained for 10,000. Changes in parameters are explicitly noted. During training, each batch was balanced to provide equal data from all environments, ensuring that the model receives the same proportion of samples per environment. {The loss-weighting hyperparameters were fixed and shared across all environments without tuning. The components were chosen to ensure that the actor and critic losses are on a comparable scale, promoting balanced gradient flow. } 

\subsubsection{Model evaluation.} We assess the model's performance on all tasks within the training distribution for the SMACv2 and GRF domains by running 500 evaluation episodes per task and computing the average win rate. For the pathfinding problem, evaluation is conducted using the POGEMA benchmark on different map sets: random maps and mazes for training tasks, and Warehouse and Cities-tiles for unseen tasks. Performance in POGEMA is measured using the average throughput metric, defined as the ratio of the total number of goals achieved by all agents to the episode length.

\subsection{Experimental Results}

\subsubsection{Multi-task MARL}
Table~\ref{tab:results-multitask} shows the complete experimental results across all three environments. In these experiments, we compare MARL-GPT (trained jointly on all environments) to five standard offline RL baselines (each trained on all tasks within a single environment). The evaluated baselines include two \textbf{memory-free} methods: Behavior Cloning (BC) and Conservative Q-Learning (CQL;~\citet{kumar2020conservative}); three \textbf{memory-based} methods: Decision Transformer (DT;~\citet{chen2021decision}), Recurrent Action Transformer with Memory (RATE;~\citet{cherepanov2026recurrent}) and a recurrent BC variant with a Long Short-Term Memory~\citep{hochreiter1997long} backbone (BC-LSTM). Crucially, all offline RL baselines are trained on the same raw observation vectors without the positional encodings: attribute, agent-index, team-index, and timestep that MARL-GPT injects into every token (see~\ref{fig:pos-example}).

\subsubsection{Expert quality.}
For SMACv2 and GRF, the expert policy is a learned model that is not near-optimal. In contrast, for POGEMA, the expert is a centralized planner (RHCR) that leverages full environment observability and a heavy search to resolve conflicts, providing strong demonstrations that serve as a useful upper bound. We aim to study the components of our model in settings where the expert offers such high-quality supervision.

Table~\ref{tab:lmapf} presents results for RHCR, MAPF-GPT (a transformer model trained with behavior cloning on MAPF tasks), MARL-GPT, and MARL-GPT-BC (behavior cloning only). In this setting, behavior cloning variants outperform MARL-GPT. This highlights that MARL-GPT's reward-based loss components may introduce instability when learning from strong expert data, and that designing effective reward functions in multi-agent pathfinding remains challenging due to the need to capture coordination, conflict resolution, and long-term planning. As a result, imitation-based approaches can more effectively leverage high-quality demonstrations like those provided by RHCR.

\begin{table}[hbt!]
    \centering
    \resizebox{\columnwidth}{!}{
    \begin{tabular}{lcccc}
        \toprule
        Scenario        & MARL-GPT & MARL-GPT-BC & MAPF-GPT  & RHCR  \\
        \midrule
        Random          & 1.16     & 1.46        & 1.50      &  2.16\\
        Mazes           & 0.96     & 1.00        & 1.09      &  1.55\\
        Warehouse       & 1.02     & 1.47        & 1.27      &  2.35 \\
        Cities-tiles    & 2.72     & 2.71        & 2.99      &  3.48 \\
        \bottomrule
    \end{tabular}
    }
    \caption{Average throughput. Comparing MARL-GPT with behavior cloning variants (MARL-GPT-BC, MAPF).}
    \label{tab:lmapf}
    \vspace{-20px}
\end{table}

\subsubsection{Unseen tasks.}
For the SMACv2 environment, we evaluated the model on previously unseen tasks with 10vs10 and 7vs7 agents, which are challenging for models trained only on 5 agents. To address this, we collected additional datasets with 10 agents and conducted two experiments: (1) We fine-tuned the pre-trained model on a small (9M) dataset with 10 agents across all races for 2,000 training steps; (2) we trained a new model from scratch using all 5-agent tasks plus an additional terran\_10\_vs\_10 task (for 3,000 and 10,000 training steps). 

The results (Table~\ref{tab:gen-smac}) show that the model can generalize to new maps with more agents if the training data include tasks with similar agent counts. Moreover, fine-tuning {with dataset} the pre-trained model requires less data and fewer steps to adapt to a new map than training from scratch.
{Two experiments show better generalization to unseen tasks. For example, (3) a model trained without terran data still performs well on new terran 5vs5 and 5vs6 scenarios. Additionally, training the model without BC loss across all races (4) can improve performance on tasks with more agents. However, including BC loss helps the model achieve better results on the training tasks but reduces its ability to generalize to new scenarios.}
All experiments were conducted with history length 4.

\begin{table}[hbt!]
    \centering
    \resizebox{\columnwidth}{!}{
    \small
    \begin{tabular}{lcccccc}
        \toprule
        \multirow{3}{*}{\shortstack{Method / \\ Environment}} & \multicolumn{2}{c}{From scratch (2)} & \multicolumn{3}{c}{Zero-shot} & \multicolumn{1}{c}{Fine-tune}\\
        \cmidrule(lr){2-3} \cmidrule(lr){4-6}
                                 & 3k     & 10k    & MARL-GPT & {w/o BC (4)}  & {w/o terran (3)} & 2k (1)  \\
        \midrule
        terran 5\_vs\_5          & 87     & 91     & 88                   & {80}                &  \heat{20}{{44*}}      & 92   \\ 
        terran 5\_vs\_6          & 48     & 57     & 52                   & {41}                & \heat{20}{{16*}}       & 62   \\ 
        terran 10\_vs\_10        & 71     & 83     & \heat{10}{0*}        & \heat{20}{{47*}}    &  \heat{10}{{0*}}       & 88   \\ 
        \midrule
        protoss 10\_vs\_10       & \heat{30}{62*}    & \heat{30}{61*}    & \heat{10}{0* }       & \heat{20}{{8*} }    &  \heat{10}{{0*} }      & 81   \\ 
        zerg 10\_vs\_10          & \heat{30}{29*}    & \heat{20}{22*}    & \heat{10}{1* }       & \heat{30}{{28*}}     & \heat{10}{ {0*}}       & 43   \\ 
        terran 7\_vs\_7          & \heat{30}{80*}    & \heat{30}{86*}    & \heat{10}{16*}       & \heat{20}{{60*}}    &  \heat{10}{{2*} }      & \heat{30}{88* } \\ 
        \bottomrule
    \end{tabular}
    } 
    \caption{Win rates \%. Comparing variants of methods on unseen tasks: models trained from scratch and pretrained (zero-shot and fine-tuned). {Each experiment used different datasets but always included tasks with zerg and protoss races (5vs5 and 5vs6). Tasks excluded from training and tested zero-shot are highlighted and marked with \textbf{*} next to their success rate. The best zero-shot results are highlighted with a stronger emphasis.}}
    \label{tab:gen-smac}
    \vspace{-20px}
\end{table}

\begin{figure}[ht!]
    \begin{center}
        \begin{minipage}[h]{0.44\linewidth}
            \includegraphics[width=1\linewidth]{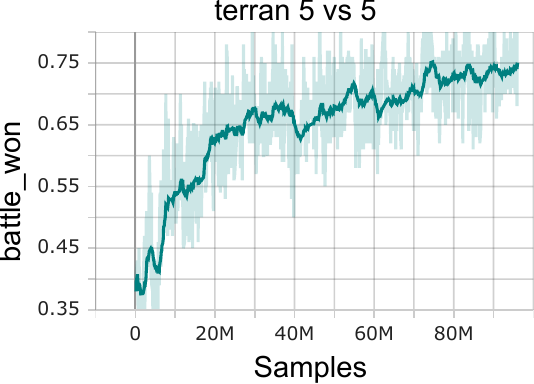} \begin{center} \textbf{(a)} \end{center} 
        \end{minipage}
        \hfill
        \begin{minipage}[h]{0.44\linewidth}
            \includegraphics[width=1\linewidth]{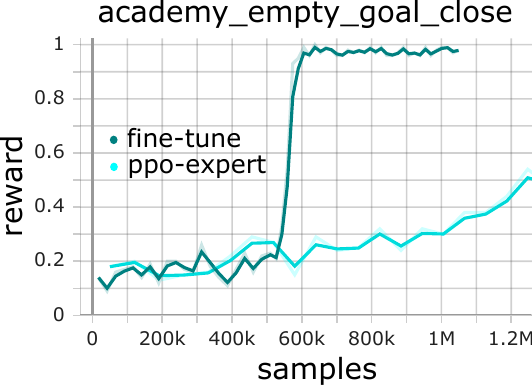} \begin{center} \textbf{\textbf{(b)}} \end{center} 
        \end{minipage}
         \hfill
        \begin{minipage}[h]{0.44\linewidth}
            \includegraphics[width=1\linewidth]{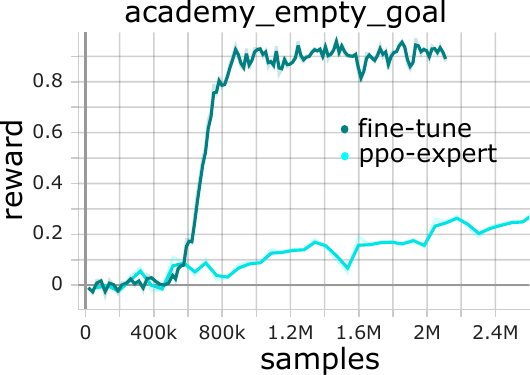} \begin{center} \textbf{(c)} \end{center} 
        \end{minipage}
        \hfill
        \begin{minipage}[h]{0.44\linewidth}
            \includegraphics[width=1\linewidth]{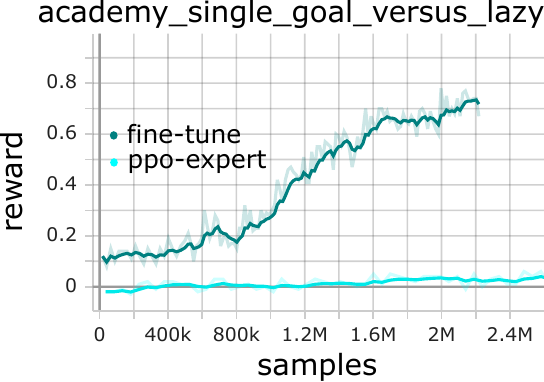} \begin{center} \textbf{\textbf{(d)}} \end{center} 
        \end{minipage}
    \end{center} 
    \caption{{Online fine-tuning results:
(a) Battle won on the terran 5vs5 task (SMACv2).
(b–d) Comparison of fine-tuned MARL-GPT and an expert model trained from scratch on GRF.}}
    \Description{Online fine-tuning results:
(a) Battle won on the terran 5vs5 task (SMACv2).
(b–d) Comparison of fine-tuned MARL-GPT and an expert model trained from scratch on GRF.}
    \label{pic:fine-tune}
    \vspace{-10px}
\end{figure}

\subsubsection{{Online fine-tune.}}
{Online fine-tune algorithm was tested on unseen tasks, in which MARL-GPT in zero-shot format achieved non-zero results. In SMACv2, we tested online fine-tuning on a new race: the initial weights were trained without the terran dataset (w/o terran in Table~\ref{tab:gen-smac}). Pretraining of the critic used 2M samples. The target task was terran 5\_vs\_5. At the beginning, the success rate (SR) was 0.4, and it improved to 0.8 by the end (Fig.~\ref{pic:fine-tune}a). For GRF, we tested on several unseen tasks (academy\_empty\_goal\_close, academy\_empty\_goal and academy\_single \_goal\_versus\_lazy) comparing results with an expert trained from scratch. Pretraining of the critic used 500k samples. Fine-tuning the pretrained model on the GRF environment showed faster adaptation to new tasks compared to training a small model from scratch (Fig.~\ref{pic:fine-tune}).}

\subsubsection{Ablation study.}
We investigated how different parameters influence the final model’s performance (Table~\ref{tab:abl}). All models were trained under identical conditions, except for one varied parameter. MARL-GPT, our final model, uses 7M parameters, a long history window of 6, positional encoding, and the full dataset. Our results show that model size, history length (notably for GRF), training data size, and positional encoding have a substantial impact on performance.

\begin{table}[hbt!]
    \centering
    \resizebox{\columnwidth}{!}{
    \begin{tabular}{lcccccccc}
        \toprule
        & No & \multicolumn{2}{c}{Model Size} & \multicolumn{2}{c}{Dataset} & \multicolumn{2}{c}{History Length} & {$\gamma$} \\
        \cmidrule(lr){3-4} \cmidrule(lr){5-6} \cmidrule(lr){7-8}
        Task & Pos. Enc. & 2M & 7M & Half & Small & 2 & 4 & {0.99}\\
        \cmidrule(lr){1-9}
        terran 5\_vs\_6       & 41 & 37  & 53  & 52 & 56 & 49 & 50 & {39}\\
        zerg 5\_vs\_6         & 33 & 36  & 40  & 38 & 40 & 37 & 39  & {29}\\
        corner         & 29 & 30  & 30  & 32 & 42 & 0  & 32 & {29} \\
        counterattack   & 88 & 85  & 88  & 87 & 20 & 20 & 0 & {85} \\
        maze         & 0.56 & 0.38 & 0.75 & 0.70 & 0.68 & 0.63 & 0.66 & {0.22}  \\
        \bottomrule
    \end{tabular}}
    \caption{
        Ablation study across tasks from SMACv2, GRF, and POGEMA. Reported are win rates (SMACv2 and GRF) and average throughput (POGEMA). We compare MARL-GPT (7M parameters) against a smaller 2M variant, and evaluate the impact of removing positional encoding, reducing dataset size, and varying history length.
    }
    \label{tab:abl}
    \vspace{-20px}
\end{table}

\subsubsection{Real-World Mini Experiment.}
We conducted a real-robot MAPF experiment; see Appendix D for details.

\subsection{Implementation Details}

The open-source repository\footnote{MARL-GPT: \url{https://github.com/Cognitive-AI-Systems/marl-gpt}}  provides all resources required to reproduce the MARL-GPT experiments, including scripts for offline training, online fine-tuning, and evaluation. 

The offline training MARL-GPT code is based on NanoGPT\footnote{NanoGPT: https://github.com/karpathy/nanoGPT} framework. This framework was selected for its straightforward and modular design, allowing for easy modification and adaptation. Training the 7M-parameter model in the main experiment for 500K iterations took 161 hours using 8 NVIDIA H100 80GB GPUs. In other experiments, training the same model for 10K iterations took 13 hours on 2 NVIDIA H100 80GB GPUs. Table~\ref{tab:model-parametrs} provides the list of hyperparameters used in our main experiments. When the model was fine-tuned with 10-agent SMACv2 data (Chapter 5.2.3) mixed with batches of original tasks to prevent catastrophic forgetting, no performance drop occurred.

\begin{table}
    \centering
    \begin{tabular}{@{}ll@{}}
        \toprule
        Parameter & Value \\
        \midrule
        Total training iterations  & 500{,}000\\
         (main experiment) & \\
        Total training iterations & 10{,}000\\
         (small experiment) & \\
        Batch size & 900\\ 
        Max size of observation & 700 \\
        Action size & 20 \\
        Number critic bins for each action & 20\\ 
        Max value & 5\\ 
        for critic regularization & \\
        $\alpha_{td}$ & 1\\
        $\alpha_{cr}$ & 1\\
        $\alpha_{a}$ & 0.1\\
        $\alpha_{bc}$ & 1\\
        history length (main experiment)  & 6\\
        history length (small experiments) & 4\\ 
        Value size per one action & 20 \\
        \midrule
        $\gamma$ & 0.95\\ 
        Target Network Update & 0.7\\ 
        Minimum learning rate & 6e-5 \\
        Maximum learning rate & 6e-4 \\
        Learning rate decay & cosine\\
        AdamW optimizer beta1 & 0.9 \\
        AdamW optimizer beta2 & 0.95 \\
        Gradient clipping & 1.0 \\
        Weight decay & 1e-1 \\
        Warm-up iterations & 2000 \\
        Use PyTorch 2.0 compilation & True \\
        Gradient accumulation steps & 16 \\
        Number of layers & 8 \\ 
        Number of attention heads & 8\\
        Embedding size & 256\\
        \bottomrule
    \end{tabular}
    \caption{Learning Hyperparameter Details.}
    \label{tab:model-parametrs}
\end{table}

\section{Conclusion}

We presented MARL-GPT, a unified transformer-based model for multi-agent reinforcement learning that operates across diverse environments using a single architecture. Trained purely from expert trajectories via imitation learning and RL, MARL-GPT achieves competitive or superior performance compared to specialized baselines in SMACv2, GRF, and POGEMA. Our results demonstrate the viability of a generalist MARL model, suggesting a path forward toward scalable, foundation models for multi-agent decision-making.

\bibliographystyle{ACM-Reference-Format} 
\bibliography{bib}

\clearpage

\section{Appendix}

\subsection{Appendix A -- Adapt Method to New Environment}

To use this model in any environment, collect expert data (e.g., using IPPO\footnote{https://github.com/alex-petrenko/sample-factory} for a specific task) containing trajectories with observation, action, and reward information. The observation is expected to be a vector. In vector environments, the meaning of the value of each element is usually known. 

Manually create a positional encoder for the environment. For each observation element obs[i], assign positional numbers as follows:
\begin{itemize}
    \item time[i]: indicates the time step (0 = current, 1 = previous, etc.). For example, if the observation includes the last five actions of one agent, these elements have time encodings 0 to 4, while agent, group, and attribute encodings remain the same.
    \item agent[i]: specifies the agent related to the information. For example, agents can be numerated (1, 2, \dots) in the observation, and information assigned to the main agent or environment can be numerated by 0.
    \item group[i]: indicates the agent’s group (0 if there is only one group). For instance, groups can represent allies (0) and enemies (1).
    \item identifies the feature type; equivalent features across agents or time share the same attribute index, otherwise they are numbered sequentially (0, 1, 2, \dots). For example, observations may include the health points (hp) of each visible agent, and these elements correspond to different agents but share the same attribute index.
\end{itemize}

The final step is to adapt the general dataloader to the new environment using the created positional encoder.

\subsection{Appendix B -- Implementation Details}

\subsubsection{Actor loss.}
When the model is trained without the behavioral cloning (BC) loss, the actor learning-rate coefficient $\alpha_a$ is reduced from $0.1$ to approximately $1/30$ in order to stabilize training in the absence of direct imitation signals.

In addition, we consider an alternative actor-loss design inspired by the Amago-2 framework~\cite{grigsby2024amago}. In this variant, the policy-gradient update is modulated by a \emph{binary} advantage filter. Concretely, we define
\begin{equation*}
f(o_t, a_t) = \mathbf{1} \{A(o_t, a_t) > 0\},
\end{equation*}
where $A(o_t, a_t)$ is the estimated advantage of action $a_t$ under observation $o_t$. The actor loss then only reinforces actions with positive advantage, making the update independent of the magnitude of the Q-values and less sensitive to the overall return scale, as in the Amago-2 actor-loss formulation. The performance with this actor loss is nearly the same as before, but choosing \(\alpha_a\) is simpler with this formulation.

\subsubsection{Action Sampling.}
The agent can select actions from the model in several ways. The model produces a probability distribution over the discrete action set, after which invalid actions are masked. Given this filtered distribution, the agent  selects the action with the highest probability (greedy selection) or samples an action according to the probabilities (stochastic sampling).

In the POGEMA environment, stochastic sampling yields the strongest performance, as multiple actions frequently receive almost identical probabilities, and randomization helps to resolve the conflict. In contrast, in SMACv2 and GRF, greedy selection is empirically more effective. Finally, actions may also be derived from the critic by interpreting critic values as preferences and inducing a sampling distribution from them.

\subsubsection{Online Fine-tuning.} We do not directly reuse the critic trained on the offline stage during online fine-tuning for two main reasons. First, in the offline stage we learn action-value functions $Q(o_t, a)$, while in the online stage we adopt a PPO-style objective that requires a state-value function $V(o_t)$ instead. Second, the offline critic is deliberately conservative and tends to severely underestimate the value of actions that were rarely or never seen in the offline dataset, which can hinder exploration during online learning. For these reasons, we train a new critic in the initial steps of the online stage.

The offline critic could be continually updated with online trajectories during the offline stage, thereby avoiding full reinitialization later. However, this would require aggregating and updating on trajectories from all environments and tasks, which is computationally expensive and significantly complicates the training pipeline.

\subsection{Appendix C -- Additional single-task comparing}
\subsubsection{Multi-agent baselines.}
We included established multi-agent online-trained baselines for SMACv2 and GRF (Table~\ref{tab:results-one-task}), specifically IPPO, QMIX, and QPLEX. These approaches use smaller networks and require less training time, but they are tailored for single tasks and some have centralized critic. Despite these advantages, MARL-GPT significantly outperformed them, consistent with previous reports~\cite{mai2024mimicking, song2023boosting}. These results demonstrate that our single model can effectively handle a variety of game instances as a unified policy.

\begin{table}[hbt!]
\centering
\resizebox{\columnwidth}{!}{
\begin{tabular}{llcccc}
\toprule
\textbf{Env} & \textbf{Task} & \textbf{IPPO} & \textbf{QMIX} & \textbf{QPLEX} &  \textbf{MARL-GPT} \\
\midrule
\multirow{3}{*}{SMAC} 
& protoss 5\_vs\_5   & 54.6 & 70.2 & 53.3 & 89 \\
& terran  5\_vs\_5   & 56.2 & 58.4 & 70.0 & 93 \\
& zerg    5\_vs\_5   & 37.2 & 37.2 & 47.8 & 74 \\
\midrule
\multirow{3}{*}{GRF} 
& pass and shoot                & 93.7 &  95.6 & 88.1 & 96  \\ 
& corner                & 50   &  20   & 28.8 & 83  \\ 
& counterattack easy    & 84.1 & 16.0  & 94.9 & 49  \\
\bottomrule
\end{tabular}
}
\caption{Win rates (\%) for selected methods. Comparing MARL-GPT and methods, which was trained just on one task.}
\label{tab:results-one-task}
\end{table}

\subsubsection{Insights from Single-Task DT}
To isolate the contribution of each SMAC scenario to cross-task generalization, we trained six single-task DT agents--one per scenario--using only 2\% of the available trajectories. This deliberately light-training regime seeks to \emph{reveal} how the source scenario shapes transfer, rather than to chase the highest possible win rates.  We then evaluated every agent on the full $6\times6$ grid of tasks; the results are summarized in Table~\ref{tab:transfer_smac} and expose two salient patterns.

First, inter-race transfer is asymmetric: a DT trained on Terran scenarios generalizes surprisingly well to Zerg tasks (e.g., 38.8 \% on $\text{terran\_5\_vs\_6} \rightarrow \text{terran\_5\_vs\_6}$ and 44.4 \% on $\text{terran\_5\_vs\_6} \rightarrow \text{zerg\_5\_vs\_5}$), while the converse holds for a Zerg-trained agent; by contrast, a Protoss-only agent shows almost no transfer beyond its own race. 
This suggests that Terran and Zerg share tactical motifs, such as kiting and surround mechanics~\citep{smacv2}, that are largely absent from Protoss micro, whose reliance on shield management and ranged focus fire produces a more domain-specific policy space. 
Second, curriculum matters: models trained on the harder 5 vs 6 variant outperform those trained on 5 vs 5 when validated on both difficulty levels of the same race, indicating that exposure to the richer state–action distribution of the difficult setting yields representations that subsume the easier case rather than overfitting to it.

\begin{table}[t]
  \centering
  \setlength{\tabcolsep}{4pt}\renewcommand{\arraystretch}{1.15}

  \resizebox{\columnwidth}{!}{%
  \begin{tabular}{lCCCCCC}
    \toprule
    \textbf{train $\backslash$ val} &
      \rot{\textbf{terran\_5\_vs\_5}} &
      \rot{\textbf{terran\_5\_vs\_6}} &
      \rot{\textbf{protoss\_5\_vs\_5}} &
      \rot{\textbf{protoss\_5\_vs\_6}} &
      \rot{\textbf{zerg\_5\_vs\_5}} &
      \rot{\textbf{zerg\_5\_vs\_6}}\\
    \midrule
    terran\_5\_vs\_5  & \heat{79}{\textbf{79.0}} & \heat{15}{15.4} & \heat{2}{2.2}  & \heat{0}{0.2} & \heat{44}{44.0} & \heat{11}{11.4}\\
    terran\_5\_vs\_6  & \heat{77}{77.0} & \heat{39}{\textbf{38.8}} & \heat{3}{3.4}  & \heat{0}{0.2} & \heat{44}{44.4} & \heat{23}{23.0}\\
    protoss\_5\_vs\_5 & \heat{1}{1.2}   & \heat{0}{0.0}  & \heat{72}{\textbf{72.4}} & \heat{6}{6.4}  & \heat{3}{2.6}  & \heat{1}{0.8}\\
    protoss\_5\_vs\_6 & \heat{0}{0.4}   & \heat{0}{0.0}  & \heat{75}{74.6} & \heat{18}{\textbf{17.6}} & \heat{1}{1.2}  & \heat{0}{0.4}\\
    zerg\_5\_vs\_5    & \heat{47}{46.8} & \heat{7}{6.8}  & \heat{3}{3.0}  & \heat{0}{0.0} & \heat{66}{\textbf{66.0}} & \heat{21}{20.8}\\
    zerg\_5\_vs\_6    & \heat{48}{47.8} & \heat{16}{16.0} & \heat{1}{0.8} & \heat{0}{0.2} & \heat{63}{63.4} & \heat{33}{\textbf{33.0}}\\
    \bottomrule
  \end{tabular}}
  \caption{{Single-Task DT on SMAC: cross-scenario validation win-rates (\%). 
  Each DT model is trained from scratch on the scenario indicated by the row and evaluated on all six scenarios listed in the columns.}}
  \label{tab:transfer_smac}
\end{table}

\subsection{Appendix D -- Robotics Experiments}

\subsubsection{Maze Environment}

\begin{figure}[ht!]     
    \centering
    \includegraphics[height=3.8cm,width=1\textwidth,keepaspectratio,page=3]{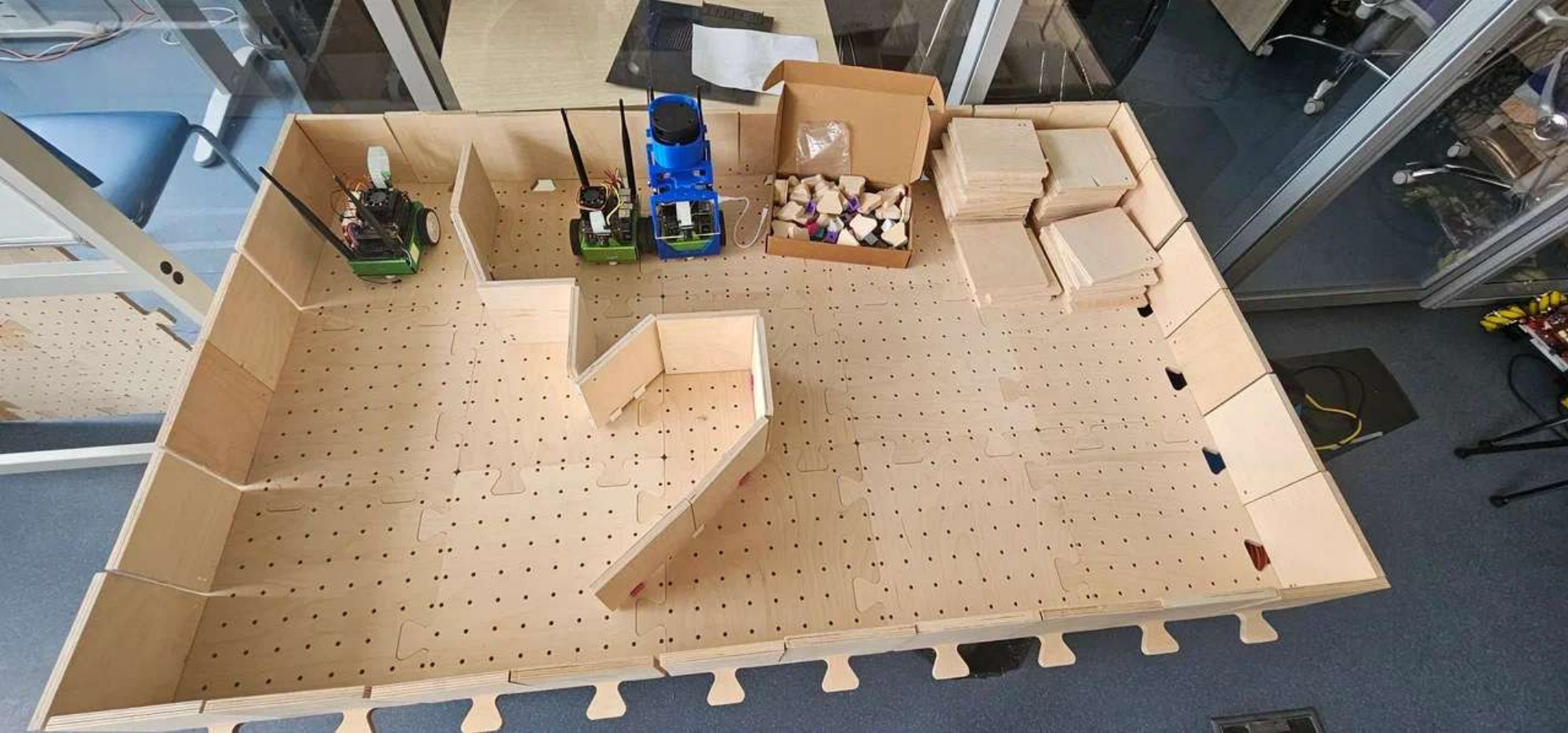}
    \caption{Modular and reconfigurable maze environment composed of standardized floor and wall units.}
    \label{fig:maze1}
\end{figure}

The physical environment is designed as a modular arena composed of standardized floor tiles and wall segments, allowing for rapid reconfiguration of the maze layout. Each floor tile is a 30~cm~$\times$~30~cm square plywood module with pre-drilled holes that support the attachment of wall elements. Wall segments are lightweight and can be mounted at any angle relative to the tile grid using custom connectors. Fig.~\ref{fig:maze1} illustrates the wall placement mechanism.

To accommodate different experimental scenarios, wall modules can be stacked to increase their height, providing visual occlusion or physical boundaries as needed. This design enables a wide range of maze configurations to be assembled with minimal effort.

\subsubsection{Robotic Agents}

Each agent is a custom-modified variant of the Waveshare JetBot platform, significantly enhanced to support onboard perception, localization, and control. The robot is equipped with a pair of DC motors with encoders and an RPLIDAR A1 2D laser scanner mounted on top for mapping and obstacle detection. A Jetson Nano 4,GB module running Ubuntu~20.04 serves as the primary onboard compute unit.

\begin{figure}[htb!]
    \centering
    \includegraphics[height=6.2cm,width=0.8\textwidth,keepaspectratio,clip,trim=0cm 7cm 0cm 5cm, page=4]{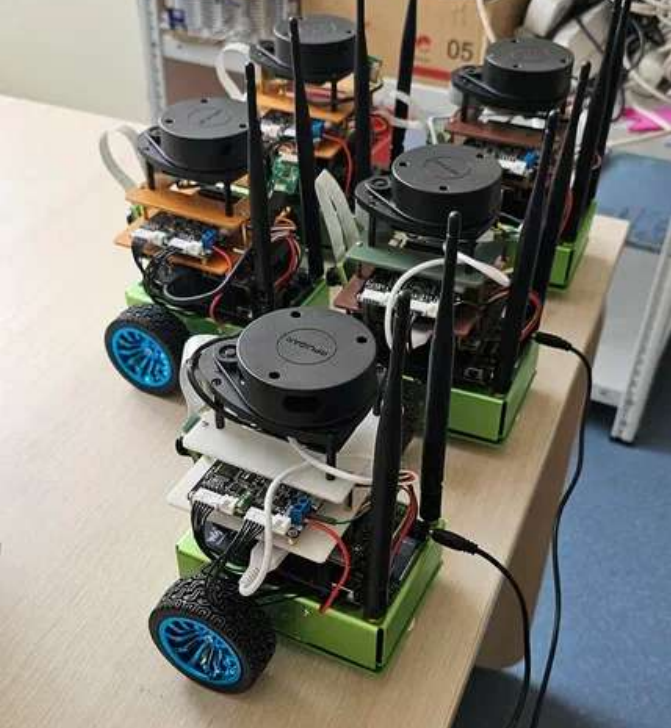}

    \caption{Waveshare JetBot robotic agent used for maze navigation experiments.}
    \label{fig:robot_agent}
\end{figure}

This is a differential-drive robot with dimensions of 17 cm × 17 cm × 22 cm (length × width × height), making it suitable for confined environments such as narrow maze corridors. Fig.~\ref{fig:robot_agent} illustrates the robots.

All software runs inside Docker containers hosted directly on the robot. The system stack includes ROS~2 Humble, compiled from source. Each robot operates independently in an isolated namespace, with segregated ROS~2 topics and TF trees, enabling scalable multi-agent deployment without inter-agent interference.

For mapping, we employ \textbf{SLAM Toolbox}, which incrementally builds a 2D occupancy grid using onboard lidar and odometry data. Navigation is handled by the \textbf{Nav2} stack, configured for both global and local planning within the constructed map. Low-level motor control is performed using the \textbf{ros2\_control} framework, integrated with a custom hardware interface developed specifically for this platform. 

\subsubsection{Experimental Protocol}

The physical maze was assembled to replicate the layout used in the corresponding simulation scenario. Due to a limited number of available wall segments, the walls were not placed in continuous contact with each other. To compensate for these small gaps, the SLAM configuration employed an adjusted \textit{inflation radius} parameter, ensuring that the robot would interpret adjacent wall segments as a continuous obstacle and avoid crossing between them. As a result, the perceived map maintained topological consistency with the simulated environment.

\begin{figure}[ht!]
    \centering
    \includegraphics[width=0.35\linewidth]{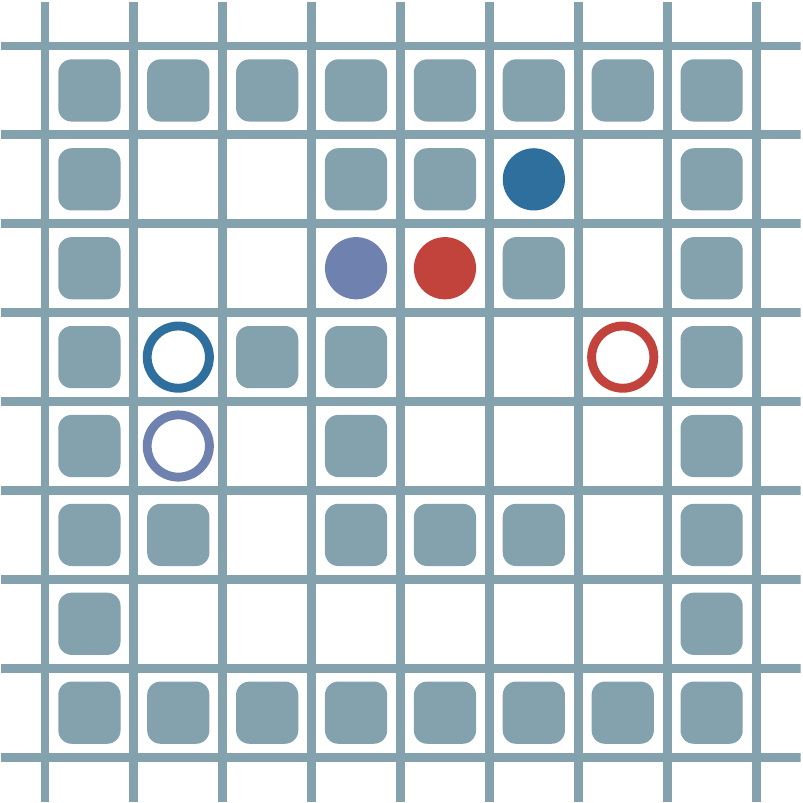}
            \hspace{10px} \includegraphics[width=0.54\linewidth,clip,trim=1cm 0cm 5cm 0cm, keepaspectratio,page=1, ]{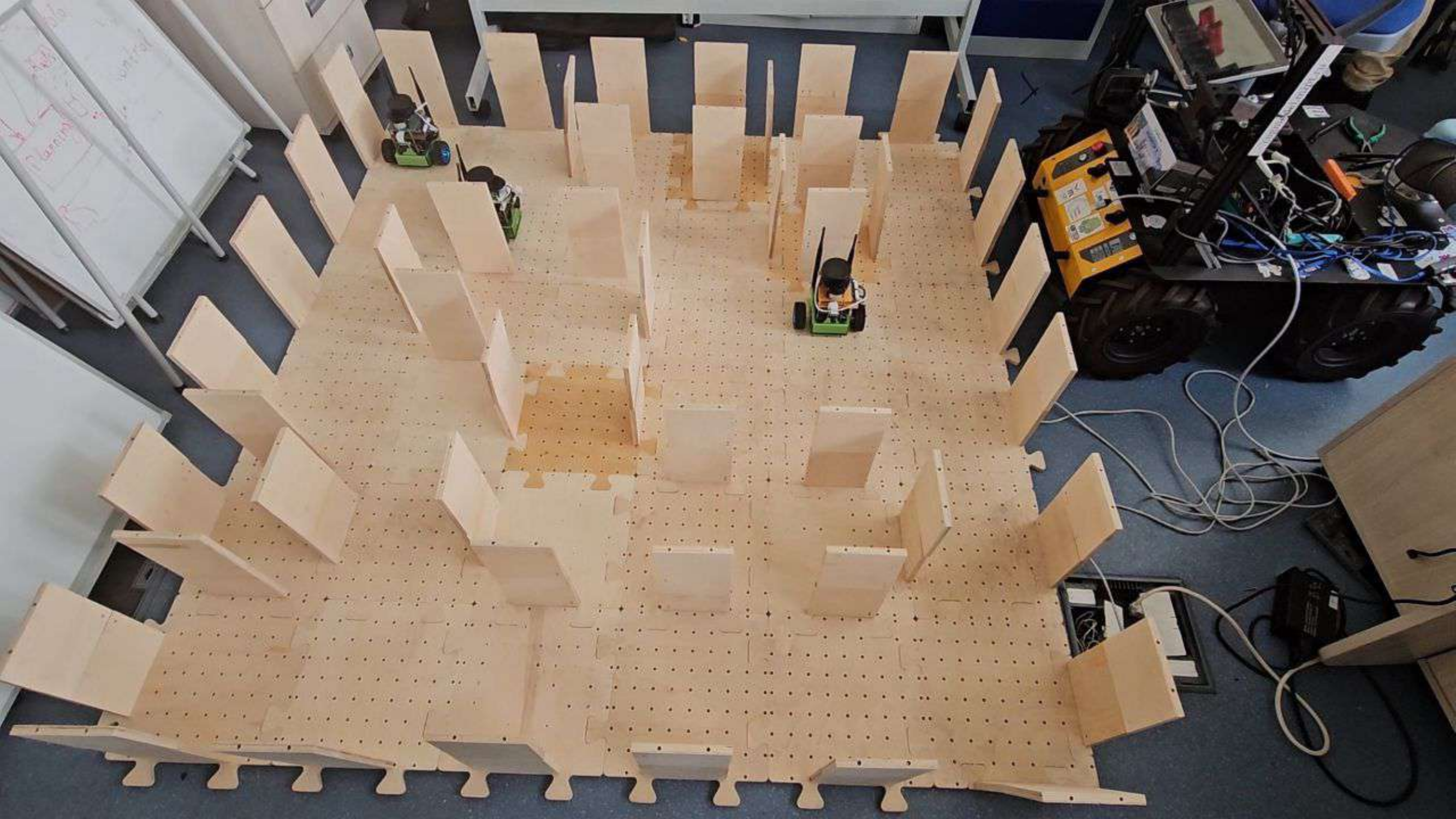}
\caption{Real-world execution of a scenario based on maze scenario with 3 robots. Left: simulated environment in POGEMA. Right: corresponding physical setup used for real-world deployment.}

    \label{fig:mazes_examples}
\end{figure}

An example of such a scenario is presented in Fig.~\ref{fig:mazes_examples}, selected from several that were tested. Shown here is an instance in the POGEMA environment alongside its corresponding real-world setup. The scenario was originally evaluated in simulation and then reconstructed in the physical environment using the same initial agent positions and target locations. The MARL-GPT policy was used without modification to control the agents during these real-world trials.

The resulting multi-agent behaviors, including coordination and conflict resolution, were recorded and are presented in the accompanying video materials. In conclusion, the experiments demonstrate that the approach is effective in the real world and that the agents are capable of coordinating their actions to successfully reach their goals.

\subsection{Generalization (limitations)}

In the long term, our goal is to enable a single foundation model to transfer cooperative multi-agent behavior across diverse environments, learning a shared representation of agent–agent interactions, and generalizing robustly to new tasks and domains. At present, this remains out of reach, as MARL-GPT still exhibits several important limitations that constrain such cross-environment transfer and zero-shot generalization.

\begin{itemize}
    \item First, our observation encoder assumes structured inputs, where each scalar feature can be augmented with positional indices (attribute type, agent id, team id, timestep). This design does not extend directly to raw visual observations, and adapting MARL-GPT to image-based input would require a  different tokenization pipeline.
    \item Second, both the observation layout and the positional index schemes are environment-specific. Each benchmark (SMACv2, GRF, POGEMA) uses its own feature set and its own rules for assigning attribute, agent, and team indices. As a result, the model has no shared latent structure that would explicitly relate semantically similar features across environments, which complicates transfer to new domains where the observation space is organized differently.
    \item Third, the action space is also bound to each environment. We rely on a shared output head with environment-dependent action masking, so the same action index can correspond to entirely different semantics in different tasks.  It hinders systematic generalization to novel action spaces and makes behavior transfer between environments non-trivial.
\end{itemize}

Consequently, MARL-GPT can be trained jointly on several diverse benchmarks and exhibits good multi-task performance within this training distribution, but its zero-shot transfer to new environments remains limited. Developing observation and action representations that are both flexible enough for heterogeneous MARL benchmarks and explicitly aligned across environments is an important direction for future work.

Despite these limitations, our experiments indicate that MARL-GPT does achieve a non-trivial degree of within-environment generalization.
\begin{itemize}
    \item In SMACv2, the model can adapt zero-shot to previously unseen tasks with different numbers of agents, provided that the training distribution includes scenarios with similar agent counts (Table 3).
    \item In POGEMA, MARL-GPT trained on random maps and mazes generalizes to new map distributions such as Warehouse and Cities-tiles (Tables 1-2). 
    \item In GRF fine-tuning lead to rapid adaptation, which we attribute to the fact that the observation space, action semantics, and agent types remain fixed across tasks (Figure 4).
\end{itemize}

Among our benchmarks, SMACv2 provides the most informative case study of the current limitations of MARL-GPT as a foundation model. Here, both the observation vectors and the action spaces change with the number and type of units, which makes systematic transfer more challenging. Manual inspection of MARL-GPT rollouts reveals two characteristic failure modes. 
\begin{itemize}
    \item First, when transferring to tasks with more agents, the model often fails to attack the last remaining enemies. In SMACv2, attack actions are indexed by enemy ID, and the set of valid indices differs between tasks. As a result, some enemy indices never appear in the training data, so the model does not learn to select the corresponding actions at test time. 
    \item Second, when training only on a subset of races, transfer to a new race can break down in the presence of unit types with no clear analog in the training distribution. For example, in the experiment where MARL-GPT is trained on zerg and protoss only, zero-shot transfer to terran5vs5 primarily fails in battles that involve healing units, whose behavior is not covered by the demonstrations (Table 3). Because fine-tuning must effectively relearn policies for such unseen unit types, adaptation in SMACv2 is noticeably slower than in GRF (Figure 4).
\end{itemize}

Taken together, these findings indicate that MARL-GPT is a first concrete step toward models that can act across several multi-agent environments, but its zero-shot abilities remain limited in terms of generalization to new environments.

\end{document}